\pdfoutput=1

\documentclass[11pt]{article}

\usepackage[]{acl}

\usepackage{times}
\usepackage{latexsym}

\usepackage{graphicx}
\usepackage{multirow}
\usepackage{array}
\usepackage{booktabs}

\usepackage{bm}
\usepackage{color,colortbl}
\usepackage{stfloats}
\usepackage{subfigure}
\usepackage{enumitem}
\usepackage{booktabs}
\usepackage{nccmath}
\usepackage{multirow}
\usepackage{subfigure}

\usepackage[T1]{fontenc}

\usepackage[utf8]{inputenc}

\usepackage{microtype}

\newcommand\blfootnote[1]{%
\begingroup
\renewcommand\thefootnote{}\footnote{#1}%
\addtocounter{footnote}{-1}%
\endgroup
}

\title{DMRST: A Joint Framework for Document-Level Multilingual \\RST Discourse Segmentation and Parsing}

\author{Zhengyuan Liu$^{\dag}$*, \ Ke Shi$^{\dag}$, \ Nancy F. Chen$^{\dag}$* \\
  Institute for Infocomm Research, A*STAR, Singapore \\
  \texttt{\{liu\_zhengyuan,shi\_ke,nfychen\}@i2r.a-star.edu.sg}}
  
\date{}

\begin{document}
\maketitle

\begin{abstract}
Text discourse parsing weighs importantly in understanding information flow and argumentative structure in natural language, making it beneficial for downstream tasks. While previous work significantly improves the performance of RST discourse parsing, they are not readily applicable to practical use cases: (1) EDU segmentation is not integrated into most existing tree parsing frameworks, thus it is not straightforward to apply such models on newly-coming data. (2) Most parsers cannot be used in multilingual scenarios, because they are developed only in English. (3) Parsers trained from single-domain treebanks do not generalize well on out-of-domain inputs. In this work, we propose a document-level multilingual RST discourse parsing framework, which conducts EDU segmentation and discourse tree parsing jointly. Moreover, we propose a cross-translation augmentation strategy to enable the framework to support multilingual parsing and improve its domain generality. Experimental results show that our model achieves state-of-the-art performance on document-level multilingual RST parsing in all sub-tasks.
\blfootnote{$^{\dag}$Equal Contribution. *Corresponding Author.}
\end{abstract}

\section{Introduction}
\label{sec:introduction}
\begin{figure}
    \begin{center}
      \includegraphics[width=0.44\textwidth]{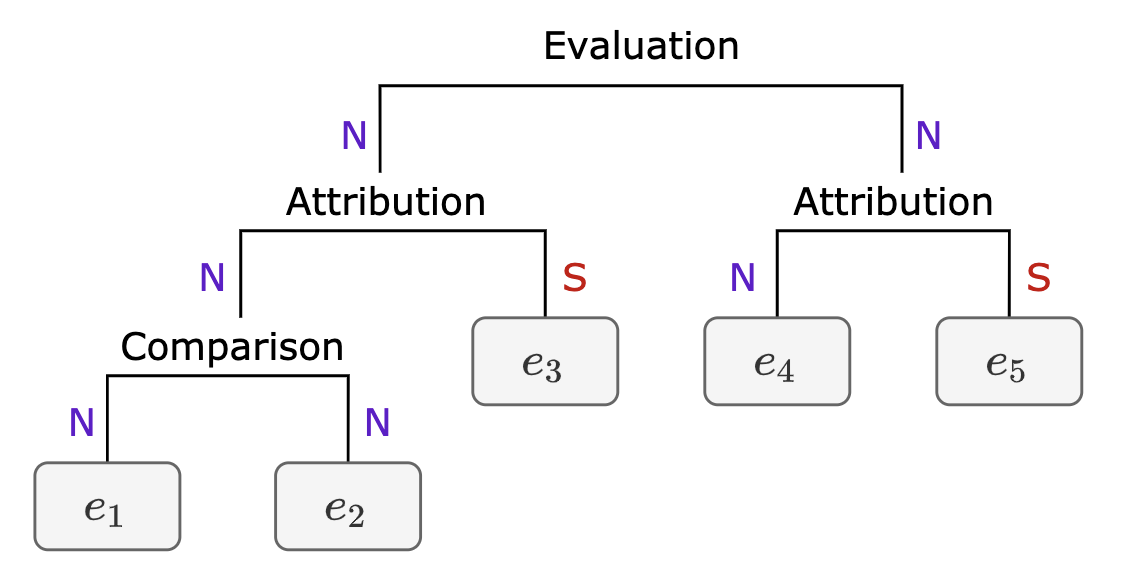}
    \end{center}
    \small{
    $e_1$$[$ The European Community's consumer price index rose a provisional 0.6\% in September from August $]$
    $e_2$$[$ and was up 5.3\% from September 1988, $]$
    $e_3$$[$ according to Eurostat, the EC's statistical agency. $]$
    $e_4$$[$ The month-to-month rise in the index was the largest since April, $]$
    $e_5$$[$ Eurostat said. $]$ 
    }
    \caption{One constituency tree with RST discourse annotation. $e_i$, $N$ and $S$ denote elementary discourse units, nucleus, and satellite, respectively. Nuclearity and discourse relations are labeled on each span pair.
    }
    \label{fig:rst_tree_example}
\vspace{-0.2cm}
\end{figure}

Rhetorical Structure Theory (RST) \cite{mann1988rhetorical} is one of the predominant theories for discourse analysis, where a document is represented by a constituency tree with discourse-related annotation. As illustrated in Figure \ref{fig:rst_tree_example}, the paragraph is split to segments named Elementary Discourse Units (EDUs), as the leaf nodes of the tree, and they are further connected by rhetorical relations (\emph{e.g.,} \textit{Elaboration}, \textit{Attribution}) to form larger text spans until the entire document is included. The spans are further categorized to \textit{Nucleus} (the core part) or \textit{Satellite} (the subordinate part) based on their relative importance in the rhetorical relations.
Thus, document-level RST discourse parsing consists of four sub-tasks: EDU segmentation, tree structure construction, nuclearity determination, and relation classification. Since discourse parsing provides structural information of the narrative flow, downstream natural language processing applications, such as reading comprehension \cite{gao2020discoReading}, sentiment analysis \cite{bhatia2015better}, and text summarization \cite{liu-chen-2019-exploiting}, can benefit from incorporating semantic-related information.

RST discourse parsing has been an active research area, especially since neural approaches and large-scale pre-trained language models were introduced. On the test set of the English RST benchmark \cite{carlson2002rst}, the performance of automatic parsing is approaching that of human annotators.
However, compared with other off-the-shelf text processing applications like machine translation, RST parsers are still not readily applicable to massive and diverse samples due to the following challenges: (1) Most  parsers take EDU segmentation as a pre-requisite data preparation step, and only conduct evaluations on samples with gold EDU segmentation. Thus it is not straightforward to utilize them to parse raw documents. (2) Parsers are primarily optimized and evaluated in English, and are not applicable on multilingual scenarios/tasks. Human annotation under the RST scheme is labor-intensive and requires specialized linguistic knowledge, resulting in a shortage of training data especially in low resource languages. (3) Data sparsity also leads to limited generalization capabilities in terms of topic domain and language variety, as the monolingual discourse treebanks usually concentrate on a specific domain. For instance, the English RST corpus is comprised of Wall Street Journal news articles, thus its parser might not perform well on scientific articles.

In this paper, to tackle the aforementioned challenges, we propose a joint framework for document-level multilingual RST discourse analysis. To achieve parsing from scratch, we enhance a top-down discourse parsing model with joint learning of EDU segmentation.
Since the well-annotated RST treebanks in different languages share the same underlying linguistic theory, data-driven approaches can benefit from joint learning on multilingual RST resources \cite{braud-etal-2017-cross-lingual}. Inspired by the success of mixed multilingual training \cite{liu2020multilingualRST}, we further propose a cross-translation data augmentation strategy to improve RST parsing in both language and domain coverage.

We conduct extensive experiments on RST treebanks from six languages: English, Spanish, Basque, German, Dutch, and Portuguese. Experimental results show that our framework achieves state-of-the-art performance in different languages and on all sub-tasks. We further investigate the model's zero-shot generalization capability, by assessing its performance via language-level cross validation. Additionally, the proposed framework can be readily extended to other languages with existing treebanks. The pre-trained model is built as an off-the-shelf application, and can be applied in an end-to-end manner.

\section{Related Work}
\label{sec:related_work}
\paragraph{RST Discourse Parsing} Discourse structures describe the organization of documents/sentences in terms of rhetorical/discourse relations. 
The Rhetorical Structure Theory (RST) \cite{mann1988rhetorical} and the Penn Discourse TreeBank (PDTB) \cite{prasad2008-pdtb} are the two most prominent theories of discourse analysis, where they are at document level and sentence level respectively. The structure-aware document analysis has shown to be useful for downstream natural language processing tasks, such as sentiment analysis \cite{bhatia2015better} and reading comprehension \cite{gao2020discoReading}.
Many studies focused on developing automatic computational solutions for discourse parsing. Statistical approaches utilized various linguistic characteristics such as $N$-gram and lexical features, syntactic and organizational features \cite{sagae2009analysis,hernault2010hilda,li2014text,heilman2015fast}, and had obtained substantial improvement on the English RST-DT benchmark \cite{carlson2002rst}. Neural networks have been making inroads into discourse analysis frameworks, such as attention-based hierarchical encoding \cite{li2016discourse} and integrating neural-based syntactic features into a transition-based parser \cite{yu2018transition}. \citet{lin2019unified} explored encoder-decoder neural architectures on sentence-level discourse analysis, with a top-down parsing procedure. Recently, pre-trained language models were introduced to document-level discourse parsing, and boosted the overall performance \cite{shi2020endRST}.

\paragraph{Multilingual Parsing}
Aside from the English treebank, datasets in other languages have also been introduced and studied, such as German \cite{stede2014-MAZ-corpus}, Dutch \cite{redeker2012-dutch-rst-dt}, and Basque \cite{iruskieta2013-basque-rst-dt}. The main challenge of multilingual discourse parsing is the sparsity of annotated data. \citet{braud-etal-2017-cross-lingual} conducted a harmonization of discourse
treebanks across annotations in different languages, and \citet{iruskieta-braud-2019-eusdisparser} used multilingual word embeddings to train systems on under-resourced languages. Recently, \citet{liu2020multilingualRST} proposed a multilingual RST parser by utilizing cross-lingual language model and EDU segment-level translation, obtaining substantial performance gains.

\paragraph{EDU Segmentation}
EDU segmentation identifies the minimal text spans to be linked by discourse relations. It is the first step in building discourse parsers, and often studied as a separated task in discourse analysis. Existing segmenters on the English discourse corpus achieve sentence-level results with 95\% F1 scores \cite{li-18-segbot}, while document-level segmentation is more challenging. \citet{muller-etal-2019-tony} proposed a discourse segmenter that supports multiple languages and schemes. Recently, taking segmentation as a sequence labeling task was shown to be effective in reaching strong segmentation results. Fusing syntactic features to language models was also introduced \cite{desai2020jointSeg}. In this work, to the best of our knowledge, we are the first to build a joint framework for document-level multilingual RST discourse analysis that supports parsing from scratch, and can be potentially extended to any language by text-level transformation.

\section{Methodology}
\label{sec:methodology}
In this section, we elaborate on the proposed joint multilingual RST discourse parsing framework. We first integrate EDU segmentation into a top-down Transformer-based neural parser, and show how to leverage dynamic loss weights to control the balance of each sub-task. We then propose cross-translation augmentation to improve the multilingual and domain generalization capability.

\subsection{Transformer-based Neural Parser}
\label{ssec:neural_parser}
The neural model consists of an EDU segmenter, a hierarchical encoder, a span splitting decoder for tree construction, and a classifier for nuclearity/relation determination.

\begin{figure*}[t!]
    \centering
    \includegraphics[width=16cm]{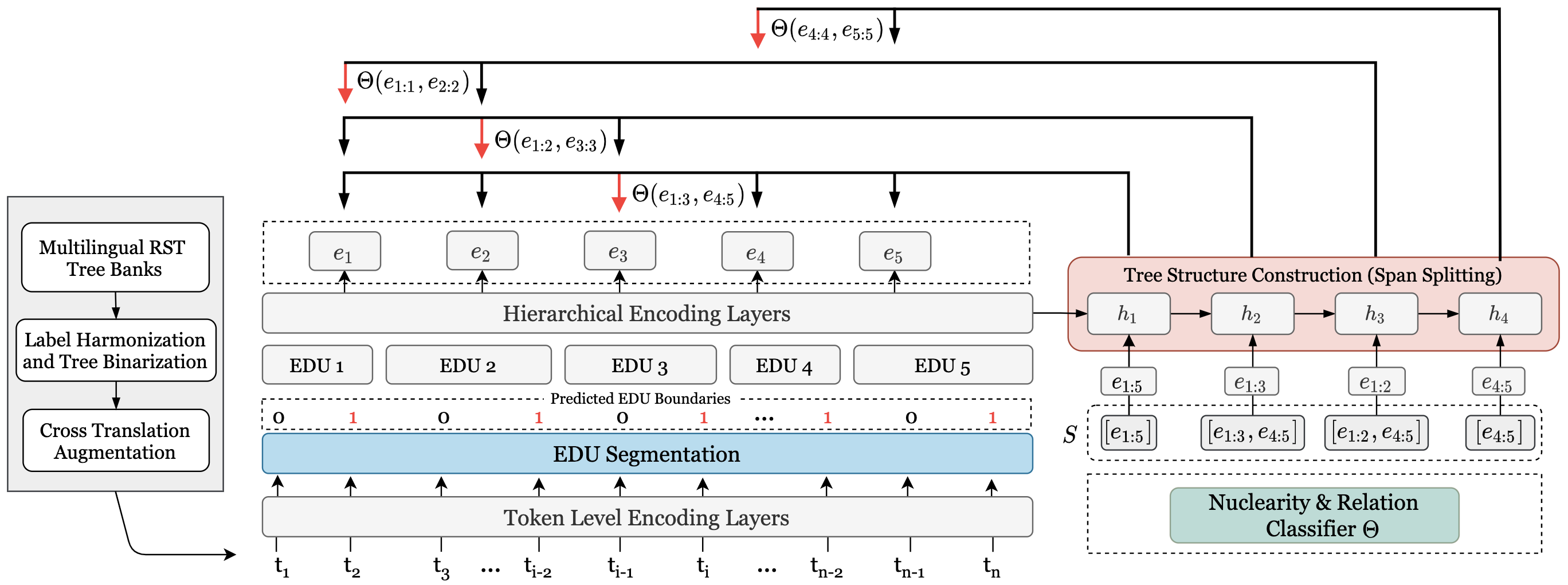}
    \caption{The architecture of the proposed joint document-level neural parser. A segmenter is first utilized to predict the EDU breaks, and a hierarchical encoder is used to generate the EDU representations. Then, the pointer-network-based decoder and the relation classifier predict the tree structure, nuclearity, and rhetorical relations. $t$, $e$ and $h$ denote input tokens, encoded EDU representations, and decoded hidden states. The stack $S$ is maintained by the decoder to track top-down depth-first span splitting. With each splitting pointer $k$, sub-spans $e_{i:k}$ and $e_{k+1:j}$ are fed to a classifier $\Phi$ for nuclearity and relation determination.}
    \label{fig:Architecture}
\end{figure*}

\subsubsection{EDU Segmentation}
The EDU segmentation aims to split a document into continuous units and is usually formulated to detect the span breaks. In this work, we conduct it as a sequence labeling task \cite{muller-etal-2019-tony,devlin-2019-bert}. Given a document containing $n$ tokens, an embedding layer is employed to generate the token-level representations $T=\{t_1,...,t_n\}$, in particular, a pre-trained language backbone is used to leverage the resourceful prior knowledge. Instead of detecting the beginning of each EDU as in previous work \cite{muller-etal-2019-tony}, here we propose to predict both EDU boundaries via token-level classification. In detail, a linear layer is used to predict the type of each token in one EDU span, i.e., at the begin/intermediate/end position.\footnote{For the EDU that only contains one token, its begin and end position are the same.}
For extensive comparison, we also implement another segmenter by using a pointer mechanism \cite{vinyals2015pointer}. Results in Table \ref{table-segmenter} show that the token-level classification approach consistently produces better performance.

\subsubsection{Hierarchical Encoding}
To obtain EDU representations with both local and global views, spans are hierarchically modeled from token and EDU-level to document-level. For the document containing $n$ tokens, the initial EDU-level representations are calculated by averaging the token embeddings $t_{i:j}$ of each EDU, where $i,j$ are its boundary indices. Then they are fed into a Bidirectional-GRU \cite{cho-etal-2014-BiGRU} to capture context-aware representations at the document level. Boundary information has been shown to be effective in previous discourse parsing studies \cite{shi2020endRST}, thus we also incorporate boundary embeddings from both ends of each EDU to implicitly exploit the syntactic features such as part-of-speech (POS) and sentential information. Then, the ensemble representations are fed to a linear layer, and we obtain the final contextualized EDU representations $E=\{e_1,...,e_m\}$, where $m$ is the total number of EDUs.

\subsubsection{Tree Structure Construction}
The constituency parsing process is to analyze the input by breaking down it into sub-spans also known as constituents. In previous studies \cite{lin2019unified,shi2020endRST}, with a generic constituency-based decoding framework, the discourse parsing results of depth-first and breadth-first manner are similar. Here the decoder builds the tree structure in a top-down depth-first manner. Starting from splitting a span with the entire document, a pointer network iteratively decides the delimitation point to divide a span into two sub-spans, until it reaches the leaf nodes with only one EDU. As the parsing example illustrated in Figure \ref{fig:Architecture}, a stack $S$ is maintained to ensure the parsing is conducted under the top-down depth-first manner, and it is initialized with the span containing all EDUs $e_{1:m}$. At each decoding step, the span $e_{i:j}$ at the head of $S$ is popped to the pointer network to decide the split point $k$ based on the attention mechanism \cite{bahdanau2014neural}.
\begin{equation}
    s_{t,u} = \sigma (h_t, e_u) \ \  \mathbf{for}\ \  u = i...j
\end{equation}
\vspace{-0.4cm}
\begin{equation}
    a_t = \mathrm{softmax}(s_t) = \frac{\mathrm{exp}(s_{t,u})}{\sum_{u=i}^j\mathrm{exp}(s_{t,u})}
\end{equation}
\noindent where $\sigma(x,y)$ is the dot product used as the attention scoring function. The span $e_{i:j}$ is split into two sub-spans $e_{i:k}$ and $e_{k+1:j}$. The sub-spans that need further processing are pushed to the top of the stack $S$ to maintain depth-first manner. The decoder iteratively parses the spans until $S$ is empty.

\subsubsection{Nuclearity and Relation Classification}
At each decoding step, a bi-affine classifier is employed to predict the nuclearity and rhetorical relations of two sub-spans $e_{i:k}$ and $e_{k+1:j}$ split by the pointer network. More specifically, the nuclearity labels \textit{Nucleus} (N) and \textit{Satellite} (S) are attached together with rhetorical relation labels (e.g., \textit{NS-Evaluation}, \textit{NN-Background}). In particular, the EDU representations are first fed to a dense layer with Exponential Linear Unit (ELU) activation for latent feature transformation, and then a bi-affine layer \cite{dozat2016deep} with softmax activation is adopted to predict the nuclearity and rhetorical relations.

\subsection{Dynamic Weighted Loss}
The training objective of our framework is to minimize the sum of the loss $\mathcal{L}_e$ of document-level EDU segmentation, the loss $\mathcal{L}_s$ of parsing the correct tree structure, and the loss $\mathcal{L}_l$ of predicting the corresponding nuclearity and relation labels: 
\begin{equation}
\small
    \mathcal{L}_e(\theta_e) = - \sum_{n=1}^N\mathrm{\log}P_{\theta_e}(y_n|X)
\end{equation}
\vspace{-0.2cm}
\begin{equation}
\small
    \mathcal{L}_s(\theta_s) = - \sum_{t=1}^T\mathrm{log}P_{\theta_s}(y_t|y_1,...,y_{t-1},X)
\end{equation}
\begin{equation}
\small
    \mathcal{L}_l(\theta_l) = - \sum_{m=1}^M\sum_{r=1}^R\mathrm{log}P_{\theta_l}(y_m=r|X)
\end{equation}
\begin{equation}
\small
    \mathcal{L}_{total}(\theta) = \lambda_1\mathcal{L}_e(\theta_e)+\lambda_2\mathcal{L}_s(\theta_s)+\lambda_3\mathcal{L}_l(\theta_l)
\end{equation}
\noindent where $X$ is the given document, $\theta_e$, $\theta_s$ and $\theta_l$ are the parameters of the EDU segmenter, the tree structure decoder, and the nuclearity-relation classifier, respectively. $N$ and $T$ are the total token number and span number. $y_1,...,y_{t-1}$ denote the sub-trees that have been generated in the previous steps. $M$ is the number of spans with at least two EDUs, and $R$ is the total number of pre-defined nuclearity-relation labels. 

To find the balance of training multiple objectives, we adopt the adaptive weighting \cite{liu2019endMultiAttn} to dynamically control the weights of multiple tasks. Specifically, each task $k$ is weighted by $\lambda_k$, where $\lambda_k$ is calculated as:
\begin{equation}
    w_k(i-1) = \frac{\mathcal{L}_k(i-1)}{\mathcal{L}_k(i-2)}
\end{equation}
\begin{equation}
\small
    \lambda_k(i) = \frac{K\cdot \mathrm{exp}(w_k(i-1)/Temp)}{\sum_j\mathrm{exp}(w_j (i-1)/Temp)}
\end{equation}
where $i$ is the training iterations, $K$ is the task number, and $Temp$ represents the temperature value that smooths the loss from re-weighting. In our experimental settings, adopting dynamic weighted loss brought about relative 2.5\% improvement on all sub-tasks.

\subsection{Cross Translation Augmentation}
\label{ssec:cross_augmentation}
Data augmentation is an effective approach to tackle the drawbacks of low resource training by creating additional data from existing samples. For instance, back translation, a popular data augmentation method, is widely applied to tasks like machine translation \cite{edunov2018backTrans}.
Since the well-annotated RST treebanks in different languages share the same underlying linguistic theory, data-driven approaches can benefit from joint learning on multilingual RST resources.
In previous work, \citet{liu2020multilingualRST} uniformed the multilingual task to a monolingual one by translating all discourse tree samples at the EDU level to English. 

\begin{figure}[t!]
    \begin{center}
      \includegraphics[width=0.45\textwidth]{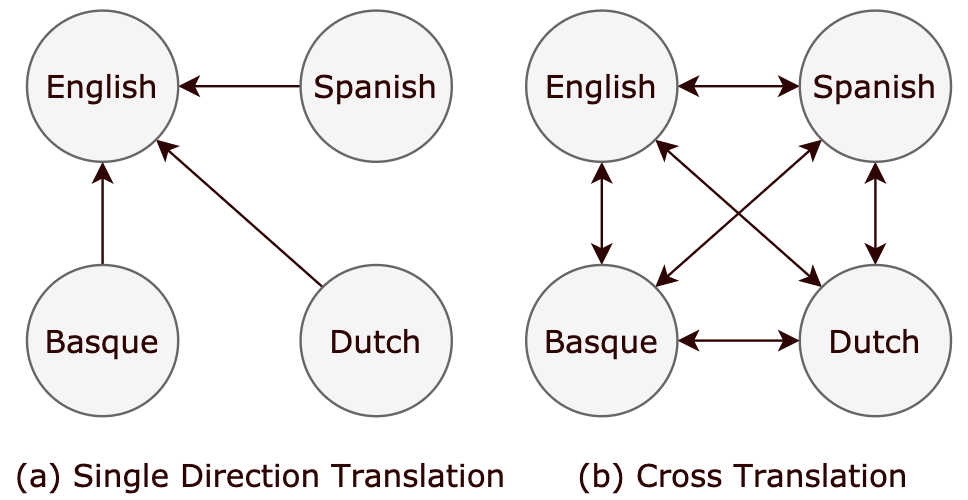}
    \end{center}
    \caption{Overview of single direction translation (a) and cross-translation strategy (b). Here we take 4 languages as an example. Arrows denote the translate directions.}
    \label{fig:cross_trans}
\end{figure}

\begin{table}[t!]
\linespread{1.0}
\footnotesize
\centering
\resizebox{\linewidth}{!}
{
\begin{tabular}{p{8cm}}
\toprule
\textbf{English (Source Text)}   \\
$e_1$$[$ The European Community's consumer price index rose a provisional 0.6\% in September from August $]$
$e_2$$[$ and was up 5.3\% from September 1988, $]$
$e_3$$[$ according to Eurostat, the EC's statistical agency. $]$
$e_4$$[$ The month-to-month rise in the index was the largest since April, $]$
$e_5$$[$ Eurostat said. $]$
\\
\midrule
\textbf{Dutch (Translated Text)}   \\
$e_1$$[$ De consumentenprijsindex van de Europese Gemeenschap is in september met een voorlopige 0,6\% gestegen ten opzichte van augustus $]$
$e_2$$[$ en steeg met 5,3\% ten opzichte van september 1988, $]$
$e_3$$[$ volgens Eurostat, het statistiekbureau van de EC. $]$
$e_4$$[$ De maand-op-maand stijging van de index was de grootste sinds april, $]$
$e_5$$[$ aldus Eurostat. $]$
\\
\midrule
\textbf{Spanish (Translated Text)}   \\
$e_1$$[$ O índice de preços ao consumidor da Comunidade Europeia subiu 0.6 \% provisório em setembro ante agosto $]$
$e_2$$[$ e aumentou 5.3 \% em relação a setembro de 1988, $]$
$e_3$$[$ de acordo com o Eurostat, a agência de estatísticas da CE. $]$
$e_4$$[$ A alta mensal do índice foi a maior desde abril, $]$
$e_5$$[$ Eurostat disse. $]$ 
\\
\bottomrule
\end{tabular}
}
\caption{One example of EDU segment-level translation. The three text samples share the same discourse tree structure, nuclearity, and relation annotation.}
\label{table-trans-example}
\vspace{-0.2cm}
\end{table}

In this paper, we propose a cross-translation data augmentation strategy.\footnote{The neural machine translation engine from Google is used: https://cloud.google.com/translate.} The method with single direction translation converts all samples to one language in both the training and the inference stage (see Figure \ref{fig:cross_trans}(a)). This approach cannot exploit the capability of multilingual language backbones. It also increases the test time due to additional computation for translation. In contrast, cross-translation will convert samples from one language to other languages, to produce multilingual training data (see Figure \ref{fig:cross_trans}(b)). Thus the model is able to process multilingual input during inference. As shown in Table \ref{table-trans-example}, adopting segment-level translation retains the original EDU segmentation as the source text, thus the converted sample in a target language will share the same discourse tree structure and nuclearity/relation labels. We postulate that this text-level transformation will bridge the gaps among different languages. Moreover, since different RST treebanks use articles from different domains \cite{liu2020multilingualRST}, we speculate that adopting cross-translation can also increase domain coverage in the monolingual space, and further improve the model's overall generalization ability.

\section{Experimental Results}
\label{sec:experiment_result}
In this section, we elaborate on experiment settings of the multilingual RST segmentation and parsing task, compare our proposed framework with previous models, and conduct result analysis.

\begin{table}[t!]
\linespread{1.0}
\centering
\resizebox{\linewidth}{!}
{
\begin{tabular}{lccc}
\toprule
\textbf{Treebank Lang.}     & \textbf{Train No.} & \textbf{Dev No.}  & \textbf{Test No.}  \\
\midrule
English (En)  &          &     &       \\
{\ } - English RST-DT  &    309      &  38   & 38 \\
{\ } - English GUM-DT  & 78 &  18   &  18      \\
Portuguese (Pt) & 256 & 38   & 38      \\
Spanish (Es) & 203    & 32    & 32   \\
German (De) & 142   & 17     & 17    \\
Dutch (Nl) & 56    &    12  & 12   \\
Basque (Eu) & 84    &  28    & 28   \\
\bottomrule
\end{tabular}
}
\caption{The collected RST discourse treebanks from 6 languages. We use the split of train, developmental and test set, as well as the data pre-processing following \cite{braud-etal-2017-cross-lingual}.}
\label{table-data}
\vspace{-0.2cm}
\end{table}

\subsection{Multilingual Dataset}
We constructed a multilingual data collection by merging RST treebanks from 6 languages: English (En) \cite{carlson2002rst}, Brazilian Portuguese (Pt)\footnote{The Portuguese RST dataset consists of 140 samples from CST-News \cite{cardoso2011cstnews}, 100 samples from CorpusTCC \cite{pardo2004-CorpusTCC}, 50 samples from Summ-it \cite{collovini2007-summ-it}, and 40 samples from Rhetalho \cite{pardo2005rhetalho}.} \cite{cardoso2011cstnews,pardo2004-CorpusTCC, collovini2007-summ-it,pardo2005rhetalho}, Spanish (Es) \cite{da2011-spanish-rst-dt}, German (De) \cite{stede2014-MAZ-corpus}, Dutch (Nl) \cite{redeker2012-dutch-rst-dt}, and Basque (Eu) \cite{iruskieta2013-basque-rst-dt}, and their details are shown in Table \ref{table-data}. We conducted label harmonization \cite{braud-etal-2017-cross-lingual} to uniform rhetorical definitions among different treebanks. The discourse trees were transformed into a binary format. Unlinked EUDs were removed. Following previous work, we reorganized the discourse relations to 18 categories, and attached the nuclearity labels (i.e., \textit{Nucleus-Satellite} (NS), \textit{Satellite-Nucleus} (SN), and \textit{Nucleus-Nucleus} (NN)) to the relation labels (e.g., \textit{Elaboration}, \textit{Attribution}). For each language, we randomly extracted a set of samples for validation. The original training size was 1.1k, and became 6.7k with cross-translation augmentation. The sub-word tokenizer of the \textit{`XLM-RoBERTa-base'} \cite{conneau-2020-XLM-Roborta} is used for input pre-processing.

\begin{table*}[t!]
\linespread{1.0}
\small
\resizebox{\linewidth}{!}
{
\begin{tabular}{lcccccc}
\toprule
& English (En)  & Portuguese (Pt) & Spanish (Es) & German (De) & Dutch (Nl) & Basque (Eu)  \\
\midrule
\citet{braud-etal-2017-cross-Seg}  & 89.5    & 82.2   & 79.3   & 85.1  & 82.6  & -    \\
\citet{muller-etal-2019-tony}  & 93.7    & 91.3   & 88.2   & 94.1  & 90.7  & 85.8    \\
Pointer-Net Segmenter  & 91.8  & 92.5  & 93.6  &  93.4 &  94.9 &  87.3  \\
Boundary CLS Segmenter (Ours)  & \textbf{96.5}  & \textbf{92.8}  & \textbf{93.7}  & \textbf{ 95.1} &  \textbf{95.5} &  \textbf{88.7}  \\
\bottomrule
\end{tabular}
}
\caption{Document-level multilingual EDU Segmentation performance on 6 languages. Micro F1 scores are reported as in \cite{muller-etal-2019-tony}.}
\label{table-segmenter}
\end{table*}

\begin{table*}[t!]
\linespread{1.0}
\small
\resizebox{\linewidth}{!}
{
\begin{tabular}{p{4.2cm}p{0.8cm}<{\centering}p{0.8cm}<{\centering}p{0.8cm}<{\centering}|p{0.8cm}<{\centering}p{0.8cm}<{\centering}p{0.8cm}<{\centering}|p{0.8cm}<{\centering}p{0.8cm}<{\centering}p{0.8cm}<{\centering}p{0.8cm}}
\toprule
& \multicolumn{3}{c}{English (En)}    & \multicolumn{3}{c}{Portuguese (Pt)} & \multicolumn{3}{c}{Spanish (Es)}   \\
Model  & Sp. & Nu. & Rel.  & Sp. & Nu. & Rel.  & Sp. & Nu. & Rel.  \\
\midrule
\citet{yu2018transition}  & 85.5    & 73.1    & 60.2                &  -  &   -  &  -  &      -     &  -  &   -   \\
\citet{iruskieta-braud-2019-eusdisparser} & 80.9    & 65.5    & 52.1     & 79.7    & 62.8    & 47.8     & 85.4    & 65.0    & 45.8     \\
Cross Rep. \cite{liu2020multilingualRST} & 87.5    & 74.7    & 63.0     & 86.3    & 71.7    & 60.0     & 86.2    & 71.1    & 54.4     \\
Segment Trans. \cite{liu2020multilingualRST}  & 87.8    & 75.4    & 63.5     & 86.5    & 72.0    & 60.3     & 87.9    & 71.4    & 56.1     \\
DMRST w/o Cross Trans.  & 87.9  &  75.3 & 64.0    &  86.5 & 73.3 &  61.5     & 88.2   & 73.7 & 60.3    \\  
DMRST (Our Framework)  & \textbf{88.2}  &  \textbf{76.2} & \textbf{64.7}    &  \textbf{87.0}  & \textbf{74.3} &  \textbf{62.1}     & \textbf{88.7}   & \textbf{75.7} & \textbf{63.4}    \\   
\midrule
& \multicolumn{3}{c}{German (De)} & \multicolumn{3}{c}{Dutch (Nl)}  & \multicolumn{3}{c}{Basque (Eu)}  \\
Model  & Sp. & Nu. & Rel.  & Sp. & Nu. & Rel.  & Sp. & Nu. & Rel.  \\
\midrule
Cross Rep. \cite{liu2020multilingualRST} & 83.6    & 62.2    & 45.1     & 85.9    & 64.5    & 49.4     & 85.1    & 65.8    & 47.7     \\
Segment Trans. \cite{liu2020multilingualRST}  & 82.3    & 58.9    & 41.0     & 84.6    & 62.7    & 47.2     & 84.4    & 65.5    & 47.3     \\
DMRST w/o Cross Trans.  & 83.1  &  62.2 & 45.9    &  85.5  & 64.4 &  50.6     & 80.2   & 59.8 & 42.1    \\  
DMRST (Our Framework)  & \textbf{84.3}    & \textbf{64.1}    & \textbf{47.3}     & \textbf{85.6}    & \textbf{66.3}    & \textbf{52.3}     & \textbf{85.1}    & \textbf{67.2}    & \textbf{48.3}    \\
\bottomrule
\end{tabular}
}
\caption{Document-level multilingual RST parsing comparison of baseline models and our framework. \textit{Sp.}, \textit{Nu.}, and \textit{Rel.} denote span splitting, nuclearity determination, and relation classification, respectively. Micro F1 scores of RST Parseval \cite{marcu2000rhetorical} are reported. Here gold EDU segmentation is used for baseline comparison.}
\label{table-result-multi}
\end{table*}

\subsection{Evaluation Metrics}
For EDU segmentation evaluation, micro-averaged F1 score of token-level segment break classification as in \cite{muller-etal-2019-tony} was used. For tree parsing evaluation, we applied the standard micro-averaged F1 scores on \textit{Span} (\textbf{Sp.}), \textit{Nuclearity-Satellite} (\textbf{Nu.}), and \textit{Rhetorical Relation} (\textbf{Rel.}), where \textit{Span} describes the accuracy of tree structure construction, \textit{Nuclearity-Satellite} and \textit{Rhetorical Relation} assesses the ability to categorize the nuclearity and the discourse relations, respectively. We also adopted \textit{Full} to evaluate the overall performance considering both \textit{Nuclearity-Satellite} and \textit{Relation} together with \textit{Span} as in \cite{morey-etal-2017-much}. Following previous studies, we adopted the same 18 relations defined in \cite{carlson2001discourse}. We reported the tree parsing scores in two metrics: the Original Parseval \citep{morey-etal-2017-much} and the RST Parseval \cite{marcu2000rhetorical} for ease of comparison with previous studies.

\subsection{Training Configuration}
The proposed framework was implemented with PyTorch \cite{paszke2019pytorch} and Hugging Face \cite{Wolf2019HuggingFacesTS}. We used \textit{`XLM-RoBERTa-base'} \cite{conneau-2020-XLM-Roborta} as the language backbone, and fine-tuned its last 8 layers during training. Documents were processed with the sub-word tokenization scheme. The dropout rate of the language backbone was set to 0.2 and that of the rest layers was 0.5. AdamW \cite{kingma2014adam} optimization algorithm was used, with the initial learning rate of 2e-5 and a linear scheduler (decay ratio=0.9). Batch size was set to 12. We trained each model for 15 epochs, and selected the best checkpoints on the validation set for evaluation. For each round of evaluation, we repeated the training 5 times with different random seeds and averaged their scores. The total trainable parameter size was 91M, where 56M parameters were from fine-tuning \textit{`XLM-RoBERTa-base'}. All experiments were run on a single Tesla A100 GPU with 40GB memory.

\begin{table*}[t!]
\linespread{1.0}
\small
\resizebox{\linewidth}{!}
{
\begin{tabular}{p{4.2cm}cccc|cccc|cccc}
\toprule   
& \multicolumn{4}{c}{English (En)}  & \multicolumn{4}{c}{Portuguese (Pt)}  & \multicolumn{4}{c}{Spanish (Es)}  \\
Model  & \multicolumn{1}{c}{Sp.} & \multicolumn{1}{c}{Nu.} & \multicolumn{1}{c}{Rel.} & \multicolumn{1}{c}{Seg.} & \multicolumn{1}{c}{Sp.} & \multicolumn{1}{c}{Nu.} & \multicolumn{1}{c}{Rel.} & \multicolumn{1}{c}{Seg.} & \multicolumn{1}{c}{Sp.} & \multicolumn{1}{c}{Nu.} & \multicolumn{1}{c}{Rel.} & \multicolumn{1}{c}{Seg.} \\
\midrule 
\multicolumn{4}{l}{Original Parseval \cite{morey-etal-2017-much}} \\
DMRST (Gold Seg.)  & 76.7   & 66.2   & 56.5 & 100.0   & 72.5   & 61.8   & 53.1  & 100.0   & 79.2  & 70.3  & 57.1 & 100.0   \\
DMRST (Predicted Seg.) & \textbf{70.4}   & \textbf{60.6}   & \textbf{51.6} & \textbf{96.5}   & \textbf{62.5}   & \textbf{51.6}   & \textbf{44.7} & \textbf{92.8} & \textbf{71.2}   & \textbf{60.1}   & \textbf{50.9} & \textbf{93.7}   \\
w/o Cross Trans. (Predicted Seg.)  & 70.3   & 60.4   & 51.3 &  96.4   & 65.3    & 53.6  & 46.3 & 93.7 & 70.2  & 59.3 & 51.1 & 93.7  \\
\midrule
\multicolumn{4}{l}{RST Parseval \cite{marcu2000rhetorical} } \\
DMRST (Gold Seg.) & 88.2   & 76.2   & 64.7 & 100.0   & 87.0   & 74.3   & 62.1 & 100.0   & 88.7   & 75.7   & 63.4 & 100.0   \\
DMRST (Predicted Seg.) & \textbf{83.2}   & \textbf{71.1}   & \textbf{60.5}  & \textbf{96.5}   & \textbf{77.8}   & \textbf{64.9}   & \textbf{53.2} & \textbf{92.8} & \textbf{79.5}   & \textbf{67.4}   & \textbf{56.7} & \textbf{93.7}   \\
w/o Cross Trans. (Predicted Seg.) & 83.0   & 70.8   & 60.7 &  96.4  & 78.4  & 65.3  & 54.7 & 93.7 & 79.4  & 66.9 & 56.5 & 93.7   \\
\midrule   
& \multicolumn{4}{c}{German (De)}  & \multicolumn{4}{c}{Dutch (Nl)}  & \multicolumn{4}{c}{Basque (Eu)}  \\
Model  & \multicolumn{1}{c}{Sp.} & \multicolumn{1}{c}{Nu.} & \multicolumn{1}{c}{Rel.} & \multicolumn{1}{c}{Seg.} & \multicolumn{1}{c}{Sp.} & \multicolumn{1}{c}{Nu.} & \multicolumn{1}{c}{Rel.} & \multicolumn{1}{c}{Seg.} & \multicolumn{1}{c}{Sp.} & \multicolumn{1}{c}{Nu.} & \multicolumn{1}{c}{Rel.} & \multicolumn{1}{c}{Seg.} \\
\midrule
\multicolumn{4}{l}{Original Parseval \cite{morey-etal-2017-much}} \\
DMRST (Gold Seg.)   & 68.6   & 45.9   & 37.1 & 100.0   & 71.2   & 54.1   & 43.1 & 100.0   & 66.6   & 48.3   & 34.7 & 100.0   \\
DMRST (Predicted Seg.) & \textbf{58.1}   & \textbf{40.1}   & \textbf{32.3} & \textbf{95.1}   & \textbf{62.3}  & \textbf{46.6}   & \textbf{39.4} & \textbf{95.5}   & \textbf{53.3}   & \textbf{39.1}   & \textbf{31.2}  & \textbf{88.7}   \\
w/o Cross Trans. (Predicted Seg.)  & 56.3   & 39.6   & 31.2 & 94.6 & 63.1   & 44.9  & 37.8 & 95.5 & 44.4  & 31.1 & 23.3 & 87.8   \\
\midrule
\multicolumn{4}{l}{RST Parseval \cite{marcu2000rhetorical} } \\
DMRST (Gold Seg.)  & 84.3   & 64.1   & 47.3 & 100.0   & 85.6   & 66.3   & 52.3 & 100.0   & 85.1   & 67.2   & 48.3  & 100.0   \\
DMRST (Predicted Seg.) & \textbf{76.4}   & \textbf{57.8}   & \textbf{41.8} & \textbf{95.1}   & \textbf{80.2}  & \textbf{62.3}   & \textbf{49.4} & \textbf{95.5}   & \textbf{71.2}   & \textbf{52.7}   & \textbf{37.2} & \textbf{88.7}  \\ 
w/o Cross Trans. (Predicted Seg.) & 75.4   & 57.0   & 41.1 & 94.6   & 80.1   & 61.9  & 48.3 & 95.5 & 66.0  & 47.5 & 33.0 & 87.8   \\
\bottomrule
\end{tabular}
}
\caption{Multilingual parsing performance comparison of using gold and predicted EDU segmentation. \textit{Sp.}, \textit{Nu.}, \textit{Rel.} and \textit{Seg.} denote span splitting, nuclearity classification, relation determination, and segmentation, respectively. Micro F1 scores of RST Parseval \cite{marcu2000rhetorical} and Original Parseval \cite{morey-etal-2017-much} are reported. Scores from the proposed framework are in bold for better readability.}
\label{table-parser-with-seg}
\end{table*}

\subsection{EDU Segmentation Results}
EDU segmentation is the first step of discourse analysis from scratch, and its accuracy is important for the follow-up parsing steps. Thus in this section, we evaluate the performance of our boundary detection segmenter, and compare it with state-of-the-art document-level multilingual EDU segmenters \cite{braud-etal-2017-cross-Seg,muller-etal-2019-tony}. Additionally, we implemented our model with a pointer mechanism \cite{vinyals2015pointer,li-18-segbot} as a control study.

From the results shown in Table \ref{table-segmenter}, our segmenter outperforms baselines significantly in all languages. This potentially results from adopting the stronger contextualized language backbone \cite{conneau-2020-XLM-Roborta}. Moreover, conducting EDU segmentation in a sequence labeling manner is more computationally efficient, and achieves higher scores than the pointer-based approach, which is consistent with the observation from a recent sentence-level study \cite{desai2020jointSeg}.

\begin{table}[t!]
    \centering
    \small
    \begin{tabular}{p{3.2cm}cccc}
    \toprule
    Model & Sp.& Nu. & Rel. & Full \\
    \midrule 
    \cite{zhang-etal-2020-top}  & 62.3   & 50.1   & 40.7 & 39.6   \\
    \cite{nguyen-etal-2021-rst}  & 68.4   & 59.1   & 47.8 & 46.6  \\
    DMRST (only EN) & 69.8 & 59.4 & 49.4 & 48.6 \\
    DMRST (Multilingual) & \textbf{70.4} & \textbf{60.6}  & \textbf{51.6} & \textbf{50.1}\\
    \bottomrule 
    \end{tabular}
    \caption{Performance comparison on the English RST treebank with predicted EDU segmentation.}
    \label{table:only_EN}
\vspace{-0.3cm}
\end{table}

\subsection{Multilingual Parsing Results}
We compare the proposed framework with several strong RST parsing baselines: \citet{yu2018transition} proposed a transition-based neural parser, obtaining competitive results in English.
\citet{iruskieta-braud-2019-eusdisparser} introduced a multilingual parser for 3 languages (English, Portuguese, and Spanish). \citet{liu2020multilingualRST} proposed a multilingual parser that utilized cross-lingual representation (\textbf{Cross Rep.}), and adopted segment-level translation (\textbf{Segment Trans.}), and produced state-of-the-art results on 6 languages. Aside from the proposed model (\textbf{DMRST}), we added an ablation study on the cross-translation strategy (\textbf{DMRST w/o Cross Trans.}). In this section, we use the gold EDU segmentation during the inference stage for a fair comparison to the baselines.

From the results shown in Table \ref{table-result-multi}:
(1) Adopting multilingual pre-trained language backbone significantly boosts the RST parsing performance.
(2) The multilingual model obtains further improvement with the cross-translation augmentation in all sub-tasks and languages.
(3) All sub-tasks are improved substantially compared to previous multilingual baselines \cite{braud-etal-2017-cross-lingual,liu2020multilingualRST}. Moreover, our model also outperforms the state-of-the-art English RST parsers (see Table \ref{table:only_EN}), demonstrating that fusing multilingual resources is beneficial for monolingual tasks.

\begin{table*}[t!]
\linespread{1.0}
\small
\resizebox{\linewidth}{!}
{
\begin{tabular}{p{3.8cm}cccc|cccc|cccc}
\toprule   
& \multicolumn{4}{c}{English (En)}  & \multicolumn{4}{c}{Portuguese (Pt)}  & \multicolumn{4}{c}{Spanish (Es)}  \\
Model  & \multicolumn{1}{c}{Sp.} & \multicolumn{1}{c}{Nu.} & \multicolumn{1}{c}{Rel.} & \multicolumn{1}{c}{Seg.} & \multicolumn{1}{c}{Sp.} & \multicolumn{1}{c}{Nu.} & \multicolumn{1}{c}{Rel.} & \multicolumn{1}{c}{Seg.} & \multicolumn{1}{c}{Sp.} & \multicolumn{1}{c}{Nu.} & \multicolumn{1}{c}{Rel.} & \multicolumn{1}{c}{Seg.} \\
\midrule 
\multicolumn{4}{l}{Original Parseval \cite{morey-etal-2017-much}} \\
DMRST w/o Cross Trans.  & 36.9  & 26.2 & 17.8 & 78.4 & 39.2 & 29.5 & 23.1 & 80.9 & 40.0 & 33.0 & 26.4 & 76.6  \\
DMRST (Our Framework) & \textbf{43.9}  & \textbf{30.8}  & \textbf{23.3} & \textbf{82.7} & \textbf{44.7} & \textbf{35.8} & \textbf{28.9} & \textbf{83.7} & \textbf{48.1} & \textbf{36.8} & \textbf{29.5} & \textbf{82.2}  \\
\midrule
\multicolumn{4}{l}{RST Parseval \cite{marcu2000rhetorical} } \\
DMRST w/o Cross Trans. & 57.8 & 40.7 & 27.0 & 78.4 & 60.4 & 44.4 & 31.8 & 80.9  & 58.1 & 42.8 & 28.3 & 76.6 \\
DMRST (Our Framework) & \textbf{63.4}  & \textbf{46.5}  & \textbf{30.2} & \textbf{82.7} & \textbf{64.5} & \textbf{50.0} & \textbf{37.7} & \textbf{83.7}  & \textbf{65.2} & \textbf{49.3} & \textbf{34.3} & \textbf{82.2}  \\
\midrule   
& \multicolumn{4}{c}{German (De)}  & \multicolumn{4}{c}{Dutch (Nl)}  & \multicolumn{4}{c}{Basque (Eu)}  \\
Model  & \multicolumn{1}{c}{Sp.} & \multicolumn{1}{c}{Nu.} & \multicolumn{1}{c}{Rel.} & \multicolumn{1}{c}{Seg.} & \multicolumn{1}{c}{Sp.} & \multicolumn{1}{c}{Nu.} & \multicolumn{1}{c}{Rel.} & \multicolumn{1}{c}{Seg.} & \multicolumn{1}{c}{Sp.} & \multicolumn{1}{c}{Nu.} & \multicolumn{1}{c}{Rel.} & \multicolumn{1}{c}{Seg.} \\
\midrule
\multicolumn{4}{l}{Original Parseval \cite{morey-etal-2017-much}} \\
DMRST w/o Cross Trans.   & 43.8 & 29.3 & 21.7 & 87.6  & 51.8 & 35.3 & 27.2 & 89.0  & 30.7 & 17.7 & 8.5 & 80.5  \\
DMRST (Our Framework) & \textbf{49.0} & \textbf{30.7} & \textbf{22.8} & \textbf{88.2}  & \textbf{56.5} & \textbf{36.0} & \textbf{27.1} & \textbf{91.0}  & \textbf{41.0} & \textbf{30.1} & \textbf{21.3} & \textbf{79.1}  \\
\midrule
\multicolumn{4}{l}{RST Parseval \cite{marcu2000rhetorical} } \\
DMRST w/o Cross Trans.  & 66.1 & 45.4 & 30.1 & 87.6  & 70.6 & 50.6 & 36.4 & 89.0  & 55.5 & 32.5 & 16.8 & 80.5  \\
DMRST (Our Framework) & \textbf{68.9} & \textbf{46.2} & \textbf{30.3} & \textbf{88.2}  & \textbf{73.9} & \textbf{52.3} & \textbf{36.1} & \textbf{91.0}  & \textbf{60.3} & \textbf{43.3} & \textbf{28.3} & \textbf{79.1}  \\
\bottomrule
\end{tabular}
}
\caption{Zero-shot performance comparison of models w/ and w/o cross-translation strategy. \textit{Sp.}, \textit{Nu.}, \textit{Rel.} and \textit{Seg.} denote span splitting, nuclearity classification, relation determination, and segmentation, respectively. Micro F1 scores of RST Parseval \cite{marcu2000rhetorical} and Original Parseval \cite{morey-etal-2017-much} are reported.}
\label{table-result-zeroshot}
\end{table*}

\subsection{Parsing from Scratch}
In most previous work on RST parsing, EDU segmentation is regarded as a separate data pre-processing step, and the test samples with gold segmentation are used for evaluation. However, in practical cases, gold EDU segmentation is unavailable. Thus in this section, we assess the proposed framework with the predicted segmentation, simulating the real-world scenario. We compare our model \textbf{DMRST} to the model without cross-translation augmentation (\textbf{DMRST w/o Cross Trans.}). Aside from the common metric RST Parseval \cite{marcu2000rhetorical} used in many prior studies, we also report test results on the Original Parseval \cite{morey-etal-2017-much}. 

From the results shown in Table \ref{table-parser-with-seg}, we observe that: (1) EDU segmentation performance of the two models are similar. This is likely because using lexical and syntactic information is sufficient to obtain a reasonable result.
(2) For both metrics, our framework achieves overall better performance in all sub-tasks and languages, especially in the lower resource languages like Basque and Dutch. (3) Since the tree structure and nuclearity/relation classification are calculated on the EDU segments, their accuracy are affected significantly by the incorrect segment predictions. For instance, when gold segmentation is provided, \textit{DMRST} outperforms \textit{DMRST w/o Cross Trans.} at all fronts. However, the former produces slightly lower scores than the latter in Portuguese, due to its suboptimal segmentation accuracy (92.8 vs. 93.7). This also emphasizes the importance of EDU segmentation in a successful end-to-end RST parsing system.

\section{Analysis on Zero-Shot Generalization}
\label{sec:zeroshot_analysis}
Incorporating discourse information is beneficial to various downstream NLP tasks, but only a small number of languages possess RST treebanks. Such treebanks have limited annotated samples, and it is difficult to extend their sample size due to annotation complexity. To examine if our proposed multilingual framework can be adopted to languages without any monolingual annotated sample  (e.g., Italian, Polish), we conducted a zero-shot analysis via language-level cross validation. 

In each round, we select one language as the target language, and RST treebanks from the remaining 5 languages are used to train the multilingual parser. We then evaluate it on the test set from the target language. For example, we assume that a small set of Portuguese articles is to be parsed, and we only have training samples from the other 5 languages (i.e., En, Es, De, Nl, and Eu). Then zero-shot inference is conducted on Portuguese.
As shown in Table \ref{table-result-zeroshot}, compared with full training (see Table \ref{table-parser-with-seg}), all the zero-shot evaluation scores drop significantly, especially on English, since the English corpus is the most resourceful and well-annotated RST treebank. Aside from English, the other 5 languages result in acceptable performance for zero-shot inference. With the cross-translation augmentation, the proposed multilingual discourse parser achieves higher scores, this is because (1) the text transformation helps language-level generalization, and (2) the mixed data have a larger domain coverage. For example, combining samples from Basque (science articles) with English (finance news) makes model perform better on Portuguese (science and news articles). This also suggests that the multilingual parser can be extended to other languages via cross-translation augmentation from existing treebanks of 6 languages.

\section{Conclusions}
In this work, we proposed a joint framework for document-level multilingual RST discourse parsing, which supports EDU segmentation as well as discourse tree parsing. Experimental results showed that the proposed framework achieves state-of-the-art performance on document-level multilingual discourse parsing on six languages in all aspects. We also demonstrated its inference capability when limited training data is available, and it can be readily extended to other languages. 

\section*{Acknowledgments}
This research was supported by funding from the Institute for Infocomm Research (I2R) under A*STAR ARES, Singapore. We thank Ai Ti Aw for the insightful discussions and Chloé Braud for sharing linguistic resources. We also thank the anonymous reviewers for their precious feedback to help improve and extend this piece of work.

\bibliography{acl}
\bibliographystyle{acl_natbib}

\newpage

\appendix


\end{document}